%% file: senina14arxiv.tex
\documentclass[10pt,twocolumn,letterpaper]{article}

\usepackage{cvpr}
\usepackage{times}
\usepackage{epsfig}
\usepackage{graphicx}
\usepackage{amsmath}
\usepackage{amssymb}
\usepackage{mathtools}

\usepackage{multirow}
\usepackage{rotating} 
\usepackage{booktabs} 

\usepackage{caption}
\usepackage{subcaption}
\usepackage{enumitem} 
\setlist{nolistsep} 

\usepackage{fancyhdr}
\usepackage{lastpage}
\usepackage{tabularx}

\usepackage[pagebackref=true,breaklinks=true,letterpaper=true,colorlinks,bookmarks=false]{hyperref}

\newcommand{\myparagraph}[1]{\vspace{-0.5cm} \paragraph{#1}}

\newcommand{\sr}[1]{\textsc{#1}}
\newcommand{\invisible}[1]{}
\graphicspath{{./fig/}{./fig/plots/}}
\newcommand{\figvspace}{\vspace{-.3cm}}
\cvprfinalcopy 

\def\cvprPaperID{456} 

\ifcvprfinal\fi
\begin{document}

\title{Coherent Multi-Sentence Video Description with Variable Level of Detail}
\newcommand{\authSpace}{&}
\author{\begin{tabular}{cccc}
Anna Senina$^{1}$ \authSpace Marcus Rohrbach$^{1}$ \authSpace Wei Qiu$^{1,2}$ \authSpace Annemarie Friedrich$^{2}$\\
Sikandar Amin$^{3}$ \authSpace Mykhaylo Andriluka$^{1}$ \authSpace Manfred Pinkal$^{2}$  \authSpace Bernt Schiele$^{1}$\\
\end{tabular}\vspace{2mm}\\
\begin{tabular}{llcrr}
\multicolumn{5}{c}{$^{1}$Max Planck Institute for Informatics, Saarbr{\"u}cken, Germany}\\
\multicolumn{5}{c}{$^{2}$Department of Computational Linguistics, Saarland University, Saarbr{\"u}cken, Germany}\\
\multicolumn{5}{c}{$^{3}$Intelligent Autonomous Systems, Technische Universit{\"a}t M{\"u}nchen, M{\"u}nchen, Germany}
\end{tabular}}

\maketitle
 
\begin{abstract}
   Humans can easily describe what they see in a coherent way and at varying level of detail.
   However, existing approaches for automatic video description are mainly focused on single sentence generation and produce descriptions at a fixed level of detail. In this paper, we
   address both of these limitations: for a variable level of detail we produce coherent multi-sentence descriptions of complex videos. We follow a two-step approach where we first learn to predict a semantic representation (SR) from video and then   
   generate natural language descriptions from the SR. To produce consistent multi-sentence descriptions, we model across-sentence consistency at the level of the SR by enforcing a consistent topic. We also contribute both to the visual recognition
   of objects proposing a hand-centric approach as well as to the robust generation of sentences using a word lattice.
   Human judges rate our multi-sentence descriptions as more readable, correct, and relevant than related work. 
   To understand the difference between more detailed and shorter descriptions, we collect and analyze a video description corpus of three levels of detail. 
       
         
\end{abstract}


\input{intro}

\input{related}

\input{dataAnalysis}
\input{approachSystem}

\input{approachMultiLevel}
\input{approachVisual}

\input{approachNLG}

\input{results}

\input{conclusion}

\small
\bibliographystyle{ieee}
\bibliography{senina14arxiv}

\end{document}

%% file: intro.tex
\section{Introduction\invisible{ - 1.25 pages}}
\label{sec:intro}

\begin{figure}[tp]
\begin{center}
\includegraphics[scale=0.2]{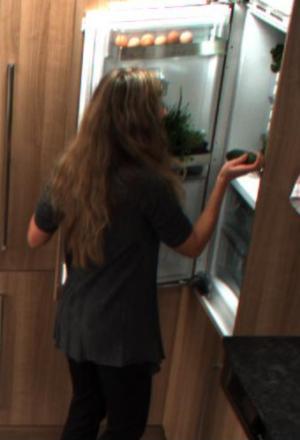}
\includegraphics[scale=0.2]{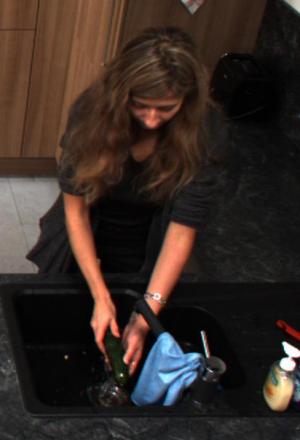}
\includegraphics[scale=0.2]{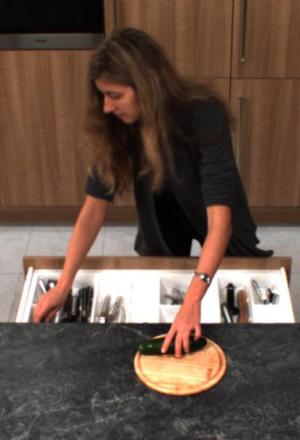}
\includegraphics[scale=0.2]{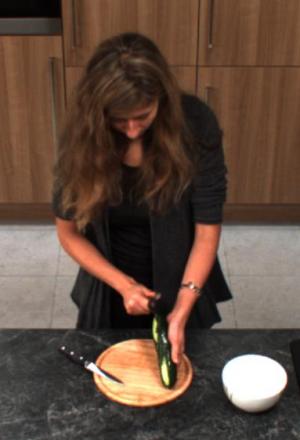}
\includegraphics[scale=0.2]{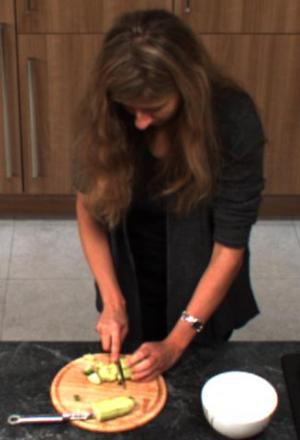}\\
\end{center}
\vspace{-0.3cm}
\footnotesize
 \begin{description}
   \item[Detailed:] 
 A woman turned on stove. Then, she took out a cucumber from
  the fridge. She washed the cucumber in the sink. She took out a cutting board and knife. She took out a plate from the drawer. She got out a plate. Next, she took out a peeler from the drawer. She peeled the skin off of the cucumber. She threw away the peels into the wastebin. The woman sliced the cucumber on the cutting board. In the end, she threw away the peels into the wastebin.
   \item[Short:] 
 A woman took out a cucumber from the refrigerator. Then, she peeled the cucumber. Finally, she sliced the cucumber on the cutting board.
   \item[One sentence:] 
  A woman entered the kitchen and sliced a cucumber.
 \end{description}
  \caption{Output of our system for a video, producing coherent multi-sentence descriptions at three levels of detail, using our automatic segmentation and extraction.}
  \figvspace
  \label{fig:teaser}
\end{figure}

Describing videos or images with natural language is an intriguing but difficult task. Recently, this task has received an increased interest both in the computer vision \cite{farhadi10eccv,kulkarni11cvpr,guadarrama13iccv,das13cvpr,rohrbach13iccv} and computational linguistic communities \cite{kuznetsova12acl,yu13acl,krishnamoorthy13aaai}.
The focus of most works on describing videos is to generate single sentences for video snippets at a fixed level of detail.
In contrast, we want to generate coherent multi-sentence descriptions for long videos with multiple activities and allow for producing descriptions at the required levels of detail (see  Figure \ref{fig:teaser}). 

The first task, multi-sentence description, has been explored for videos
 previously \cite{das13cvpr,khan11iccvw,tan11mm}, but open challenges remain, namely finding a segmentation 
 of appropriate granularity and  
generating a conceptually and linguistically coherent description.
Latter is important as changes in topic lead to unnatural descriptions.
To allow reasoning across sentences we use an intermediate semantic representation (SR) which is inferred from the video.
For generating multi-sentence descriptions we ensure that sentences describing different activities are about the same topic. Specifically, we predict the dish that is being prepared in our cooking scenario. 
We improve intra-sentence consistency
by allowing our language model to choose from a probabilistic SR rather than a single MAP estimate. Finally we apply linguistic cohesion to ensure a fluent text.  

Our second focus is generating descriptions with a varying level of detail. While this is a researched problem in natural language generation, e.g. in context of user models \cite{zukerman2001natural}, we are not aware of any work in computer vision which studies how to select the desired amount of information to be recognized. 
To understand which information is required for producing a description at a desired level of detail we collected descriptions at various levels of detail for the same video and analyzed which aspects of the video are verbalized in each case. Our analysis suggests that shorter descriptions focus on more discriminative activities/objects for a given topic. We propose to handle this by verbalizing only the most relevant video segments according to a predicted topic.
A second conclusion is that for detailed descriptions in our application domain of cooking activities, it is important to describe all handled objects, in which current approaches have only shown limited success. For this we propose a hand-centric object recognition model, that improves recognition of manipulated objects. 

The contributions of the paper are as follows.
The first main contribution is to generate coherent multi-sentence descriptions. For this we propose a model which enforces conceptual consistency across sentences (Sec. \ref{sec:approach_system}) as well as linguistic coherence (Sec. \ref{sec:approach_nlg}). 
Our second main contribution is to allow generation of descriptions at a desired level of detail. For this we collected, aligned, and analyzed a corpus of descriptions of three levels of detail
(Sec. \ref{sec:dataAnalysis}).
Based on our analysis we explore and evaluate different options to generate short video descriptions. 
Third, we significantly improve the visual recognition (Sec. \ref{sec:results}) based on our hand-centric approach (Sec. \ref{sec:approach_visual}).



%% file: related.tex
\section{Related Work\invisible{ - 0.5 pages}}
\label{sec:related}

In the following we discuss the most relevant work on image and video description with a focus on coherent multi-sentence and multi-level language generation. 
To generate descriptions for videos and images, rules or templates are a powerful tool but need to be manually defined~\cite{kulkarni11cvpr,tan11mm,gupta09cvpr,krishnamoorthy13aaai,guadarrama13iccv}. An alternative is to retrieve sentences from a training corpus~\cite{farhadi10eccv,das13cvpr} or to compose novel descriptions based on a language model~\cite{kulkarni11cvpr,kuznetsova12acl,mitchell12eacl,rohrbach13iccv}. 
We base our approach on \cite{rohrbach13iccv} which uses an intermediate SR modeled with a CRF. It uses statistical machine translation (SMT)~\cite{koehn07acl} to translate the SR to a single sentence for a manually segmented video-snippet. 
In contrast we segment the video automatically, produce multi-sentence descriptions for an entire video at multiple levels of detail. 
Furthermore, we exploit the probabilistic output of the CRF and incorporate it in the SMT using a word-lattice \cite{dyer2008generalizing}.

Multi-sentence generation has been addressed for images by combining descriptions for different detected objects. \cite{kulkarni11cvpr} connects different object detection with prepositions using a CRF and generates a sentence for each pair. \cite{kuznetsova12acl} models discourse constraints, content planning, linguistic cohesion, and is able to reduce redundancy using ILP. In contrast we model a global semantic topic to allow descriptions with many sentences  while \cite{kuznetsova12acl} generates in most cases only 1-3 sentences.

For videos, \cite{gupta09cvpr} learns AND/OR graphs to capture the causal relationships of actions given visual and textual data. During test time they find the most fitting graph to produce template-based, multi-sentence descriptions. 
\cite{khan11iccvw} produces multiple sentences and use paraphrasing and merging to get the minimum needed number of sentences. In contrast we model consistency across sentences.
Using a simple template, \cite{tan11mm} generates a sentence every 10 seconds based on concept detection.  For consistency they recognize a high level event and remove inconsistent concepts. This has similarity to our idea of a topic but they work in a much simpler setting of just 3 high level events with manually defined relations to all existing concepts.  
To generate multiple sentences for a video, \cite{das13cvpr}  segments the video based on the similarity of concept detections in neighboring frames. In contrast we use agglomarative clustering of attribute classifiers trained to capture the desired granularity. Next, \cite{das13cvpr} ensures that their low level detections are consistent with their concept prediction and retrieve the most likely training sentence. While their verbs are manually defined for all  concept pairs, we focus on activity recognition and describing activities with verbs predicted by SMT.    
While SMT has mostly focused on the translation of single sentences, recent approaches aim to optimize the entire translation in order to generate consistent text \cite{hardmeier13acl}. It would be interesting to combine this idea with the probabilistic output of our CRF as part of future work.
  

We are not aware of any work in computer vision approaching descriptions at different levels of detail. Closest is  \cite{guadarrama13iccv}, which predicts more abstract verbs and nouns if the uncertainty is too high for a more specific prediction. Our approach is complementary, as our goal is to produce  different detailed descriptions, using abstraction to summarize over multiple activities or objects, rather than to decrease uncertainty. Our work is also different from video summarization as it solves a different task, namely getting a visual summary rather than a textual description.


%
%

%% file: dataAnalysis.tex
\section{Analysis of human-written video descriptions of different levels of detail\invisible{ - 0.5 pages}}
\label{sec:dataAnalysis}

An important goal of our work is to generate natural language descriptions for videos at different levels of detail. In this section, we investigate which aspects of a video are verbalized by humans and how descriptions of different levels of detail differ, with the aim of obtaining a better understanding of what needs to be recognized in a video.


\myparagraph{Data collection}
The data was collected via Amazon Mechanical Turk (AMT) using the TACoS
corpus~\cite{regneri13tacl}. The corpus contains 127 cooking videos of 26 different dishes and aligned text descriptions. For each video  we asked a person to describe it in three ways: (1) a detailed description with at most 15 sentences, (2) a short description (3-5 sentences), and (3) a single sentence. 
Unlike~\cite{regneri13tacl}, workers could freely describe videos without aligning each sentence to the video.
Our data collection hence results in more natural descriptions, having a more complex sentence structure (e.g., they make use of temporal connectives and anaphora).
To ensure a high quality we manually excluded descriptions violating our requirements as well as irrelevant texts.
Overall, we have collected about 2600 triples of descriptions for TACoS videos.

\myparagraph{Analysis of human-written descriptions}
First, we analyze the collected descriptions with respect to which aspects of the videos are verbalized. 
We assign part-of-speech (POS) tags to the collected descriptions and the ones provided by TACoS using the Stanford POS tagger \cite{toutanova2003feature}. Any word tagged as a verb is considered to be an \sr{activity}, and any word tagged as an adjective is considered to represent an \sr{attribute}. We classify all adverbials as providing \sr{spatial} or \sr{temporal} information using a hand-compiled list of adverbials. \sr{quantity} information is assumed when one of the words has been tagged as a cardinal number or when a noun is a hyponym, i.e., in an \textit{is-a} relation, of `quantity' or `portion' in WordNet \cite{miller1995wordnet}. We use \sr{person}, \sr{food}, \sr{tool}, \sr{utensil} or \sr{appliance} and categories for nouns. To identify the category of a specific noun, we check whether the words are hyponyms of appropriate WordNet entries, and additionally check manually created white- and blacklists for each category. \sr{food} is considered to be any edible item or dish. \sr{tools} are items such as \textit{knife} or \textit{chopper}, while \sr{utensils} are other kitchen utensils such as \textit{bowl} or \textit{cutting board}. Finally, the \sr{appliance} category comprises non-movable items such as \textit{stove}, \textit{kitchen} or \textit{sink}.

\begin{figure}[t]
\center
\includegraphics[trim = 20mm 197mm 40mm 27mm, clip, width=0.95\linewidth]{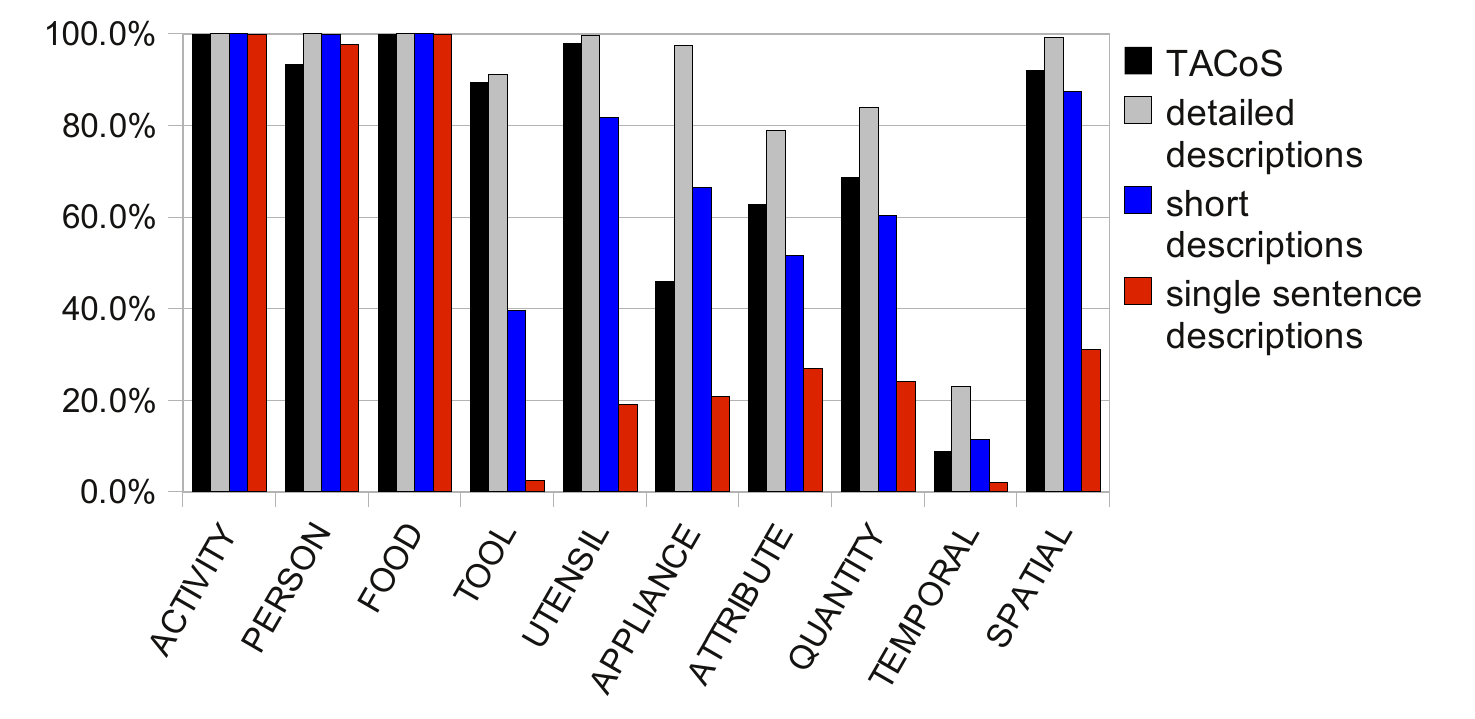}
\vspace{-.3cm}
\caption{Percentage of descriptions in which a category was verbalized.}
\label{fig:verbalization}
\figvspace
\end{figure}

Figure \ref{fig:verbalization} shows the percentages of descriptions in which at least one word of the respective category occurred. 
\sr{activities}, \sr{food} items and the \sr{person} are mentioned in almost all descriptions. 
For \sr{tools}, \sr{utensils},  \sr{appliances}, \sr{attributes},  \sr{quantities}, and \sr{spatial} the occurrence frequency decreases as the descriptions become shorter. \sr{tools}, \sr{utensils}, and \sr{appliances} nearly fully disappear  
in the single-sentence descriptions. 
The detailed descriptions and the descriptions from TACoS are similar except in the \sr{appliance} category.

%



Next, we performed a qualitative comparison of the 10  most frequent activities/food items verbalized in different types of descriptions. 
The descriptions from TACoS, the detailed descriptions and the short description
mainly use verbs describing specific activities, such as \textit{cut} or \textit{take}, see Table \ref{tab:verbs}.
In the single-sentence descriptions, verbs such as \textit{prepare}, \textit{cook} and \textit{make}, which summarize a set of activities, are frequently used. This indicates that when generating single sentence descriptions of videos, it may not be sufficient to simply extract sentences from the longer descriptions, but some degree of abstractive summarization is needed.
\input{tables-language}
We also compared most top-10 food items mentioned in the collected descriptions, see Table \ref{tab:food}. 
Due to the relative simplicity of the dishes present in TACoS (e.g. as \textit{preparing a carrot}), we do not observe much variation in the degree of abstractness of the used words. However, there is a difference in which words are verbalized. 
While the detailed descriptions frequently mention common ingredients such as \textit{water}, \textit{salt} or \textit{spice}, this is less for the short descriptions, and almost never for the single sentence descriptions.
In short descriptions humans mention the objects that are more relevant for the respective dish, which are usually the main ingredients such as \textit{potato} or \textit{carrot}, and skip the rest. Correspondingly, in single sentence descriptions humans only focus on the main ingredients. This suggests that knowing the dish that is being prepared is necessary in order to determine the important objects to be verbalized.

\myparagraph{Discussion}
We draw four conclusions from this analysis. First, in detailed descriptions all fine-grained activities and objects are mentioned. This means that the visual recognition system ideally should identify all of them. 
Second, short descriptions could be obtained from detailed descriptions using extractive summarization techniques. One might apply extractive summarization purely on the language side, but we explore an extractive technique on the visual side. However, the fact that the various levels show different relative frequency of verbalized concepts  indicates that a specific translation model targeted to desired type of descriptions might be beneficial to match the SR with the text. 
Third, single-sentence descriptions qualitatively differ from all other description types, which suggests that abstractive summarization is required for this level.
Forth, it is advantageous to explicitly model and recognize the dish that is prepared. This also helps to generate consistent multi-sentence descriptions, another important goal of this paper. 

%
%


%% file: tables-language.tex
\begin{table}
\centering
\small
\begin{tabular}{|l|l|l|l|}
\hline
TaCos & detailed & short & single sentence\\
\hline
\hline
cut	& cut & cut &	cut\\
take &	take &	take &	slice\\
get &	put &	put &	peel\\
put	& place &	get & 	chop\\
wash & 	get & 	slice &	enter\\
place &	wash &	peel &	\textbf{prepare}\\
rinse &	remove &	place &	\textbf{cook}\\
remove	& rinse & 	wash	 & dice\\
peel & 	slice &	enter &	\textbf{make}\\
be & 	throw &	walk &	walk\\
\hline
\end{tabular}
\caption{Top 10 verbs in video descriptions. Words marked with bold are the ones that only appear in single sentence descriptions.}
\label{tab:verbs}
\end{table}

\begin{table}
\centering
\small
\begin{tabular}{|l|l|l|l|}
\hline
TaCos & detailed & short & single sentence\\
\hline
\hline
\textbf{water}	& \textbf{water}	& \textbf{water}	&	orange\\
\textbf{salt}	& \textbf{salt} 	& orange 			&	egg\\
\textit{pepper}	& \textit{pepper}	& egg				&	\textit{pepper}\\
juice			& orange			& \textit{pepper}	&	juice\\
orange 			& egg				& juice				&	onion\\
egg				& \textbf{oil}		& onion				&	cauliflower\\
\textbf{oil}	& \textbf{spice}	& potato			&	pineapple\\
herb	 		& juice				& cauliflower		&	carrot\\
\textbf{butter}	& onion				& pineapple			&	broccoli\\
onion			& fruit				& carrot			&	avocado\\
\hline
\end{tabular}
\caption{Top 10 food items in video descriptions. Words marked with bold are the ones that appear in all but single sentence descriptions. Note that \textit{pepper} is an exception, as it can refer to spice as well as a vegetable, and therefore appears in all cases.}
\label{tab:food}
\figvspace
\end{table}

%% file: approachSystem.tex
\section{Generating consistent multi-sentence video descriptions at multiple levels of detail\invisible{ - 1.25 pages}}
\label{sec:approach_system}

First we present our approach to generate consistent multi-sentence descriptions for a video with a given temporal segmentation and then describe our segmentation approach.
Next, we present our approach to produce video descriptions on multiple levels on detail. We produce short and one sentence descriptions, using the obtained video segmentation by selecting the most relevant intervals given the predicted topic (dish).
\subsection{Multi-sentence video descriptions}
Assume that a video $v$ can be decomposed into a set of $I$ video snippets represented by video descriptors $\{x_1,...,x_i,...,x_I\}$, where each snippet can be described by a single sentence $z_i$. To reason across sentences we employ an intermediate semantic representation (SR)  $y_i$. 
We base our approach for a video snippet on the translation approach proposed in \cite{rohrbach13iccv}. We chose this approach as it allows to learn both the prediction of a semantic representation $x\rightarrow y$ from visual training data $(x_i,y_i)$ and the language generation $y\rightarrow z$ from an aligned sentence corpus $(y_i,z_i)$.
While this paper builds on the semantic representation from \cite{rohrbach13iccv}, our idea of consistency is applicable to other semantic representations.
The SR $y$ is a tuple of activity and participating objects/locations, e.g. in our case 
\sr{$\langle$activity, tool, object, source, target$\rangle$}. The relationship is modeled in a CRF where these entities are modeled as nodes $n\in \{1,...,N\}$ ($N=5$ in our case) observing the video snippets $x_i$ as unaries. We define $s_n$ as a state of node $n$, where $s_n\in S$. We use a fully connected graph and linear pairwise (p) an unary (u) terms.
In addition to \cite{rohrbach13iccv}, to enable a consistent prediction within a video, we introduce a high level topic node $t$ in the graph, which is also connected to all nodes. However, in contrast to the other nodes it observes the entire video $v$ to estimate its topic rather than a single video snippet. For the topic node $t$ we define a state $s_t\in T$, where $T$ is a set of all topics. We then use the following energy formulations for the structured model:
\begin{equation}
\begin{multlined}
\label{eq:crfEnergy}
E(s_1,..., s_N,s_t|x_i,v)= \\
\sum_{n=1}^N E^u(s_n|x_i)+E^u(s_t|v)+\sum_{n\sim m}E^p(s_n,s_m)
\end{multlined}
\end{equation}
with $E^p(s_n,s_m)=w^p_{n,m}$, where $w^p_{n,m}$ are the learned pairwise weights between the CRF node-state $s_n$ and node-state $s_m$. We discuss the unary features in Sec. \ref{sec:approach_visual}.

While adding the topic node makes each video snippet aware of the full video, it does not enforce consistency across snippets. 
Thus, at test time, we compute the conditional probability $p(s_1,...,s_N|\hat{s_t})$, setting $s_t$ to the highest scoring state $\hat{s_t}$ over all segments $i$:
\begin{equation}
\label{eq:topic}
\hat{s_t}=\arg\max_{s_t\in T}\max_{i \in I} p(s_t|x_i)
\end{equation}
We learn the model by independently training all video descriptors $x_i$ and SR labels $ y_i=\langle s_1,s_2, \ldots,s_N, s_t \rangle$ using loopy belief propagation implemented in \cite{schmidt13web}.
The possible states of the CRF nodes are based on the provided video segment labels for the TACoS dataset \cite{regneri13tacl} as well as the topic (dish) labels of the videos.

\myparagraph{Segmentation} 
For our above described approach, we have to split the video $v$ into video-snippets $x_i$. Two aspects are important for this temporal segmentation: it has to find the appropriate granularity so it can be described by a single sentence and it should not contain any unimportant (background) segments which would typically not be described by humans. For the first aspect, we employ agglomerative clustering on a score-vector of semantic attribute classifiers (see Sec. \ref{sec:approach_visual}). These classifiers are trained to capture the annotation granularity. We found that the raw video features are not able to capture this very well. The second aspect is achieved by training a background classifier on all unlabeled video segments as negative examples versus all labeled snippets as positive. 


%% file: approachMultiLevel.tex
\subsection{Multi-level video descriptions}


Based on the observations discussed in Sec. \ref{sec:dataAnalysis}, we propose to generate shorter descriptions by extracting a subset of segments from our segmentation. We select relevant segments by scoring how discriminative their predicted SR is for the predicted topic by summing the $tfidf$ scores of the node-states, computed on the training set. For the SR $\langle s_1,\ldots,s_N,s_t\rangle$, its score $r$ equals to:
\begin{equation}
\begin{multlined}
\label{eq:relevance}
r(s_1,...,s_N,s_t)=\sum_{n=1}^N tfidf(s_n,s_t)
\end{multlined}
\end{equation}   
where $tfidf$ is defined as the normalized frequency of the state $s_n$ (i.e. activity or object) in topic $s_t$ times the inverse frequency of its appearance in all topics: 
\begin{equation}
\textstyle 
\label{eq:tfidf}
tfidf(s_n,s_t) = \frac{f(s_n,s_t)}{\max_{s'_n\in S}f(s'_n,s_t)}\log\left(\frac{|T|}{\sum_{s'_t\in T}f(s_n,s'_t)>0}\right)
\end{equation}
This way we select the $n$ highest scoring segments and use them to produce a short description of the video. 
One way to produce a description would be to simply extract sentences that correspond to selected segments from the detailed description.
However, given that some concepts are not verbalized in shorter descriptions, as pointed out in Sec. \ref{sec:dataAnalysis}, we additionally explore the approach of learning a translation model targeted to the desired level of detail.
We similarly produce single sentence descriptions and also compare our approach to the retrieval baseline (see Sec. \ref{sec:results:multilevel}).

%% file: approachVisual.tex
\section{Improving Visual features\invisible{ - 1 page}}
\label{sec:approach_visual}

One of the conclusions drawn in \cite{rohrbach13iccv} is that the noisy visual
recognition is a main limitation of the suggested translation approach.
Therefore, we are aiming to improve the quality of predicted semantic
representations. Moreover, given that we want to infer the topic (dish)
(Sec. \ref{sec:approach_system}), it is particularly important to recognize such
challenging objects as food items.

The visual recognition approach of \cite{rohrbach13iccv} is based on
dense trajectory features \cite{wang13ijcv}. In \cite{rohrbach13iccv} the features are quantized in a codebook
and used to train the visual attribute classifiers. Finally, the classifiers' score vectors are used as features for the CRF unaries.
We improve this approach in two ways. First we change the features used
for CRF unaries to the semantic unaries. Second, in order to
improve the object recognition, we suggest that it is beneficial to focus on hands'
regions, rather than to use holistic features, such as dense trajectories. This
observation is intuitive, in particular in domains, where people mostly perform
hand-related activities. We develop a robust hand detector
and extract color Sift features in
hands' neighborhood to recognize the manipulated objects.

\subsection{Semantic unaries}
As mentioned above, the approach of \cite{rohrbach13iccv} uses visual attributes to obtain the features for CRF unaries. One problem with this approach is that it ignores the semantic meaning of the attributes. E.g. a classifier for a visual attribute \emph{knife} is learned disregarding whether a knife is a \sr{tool} (e.g. \emph{cut with a knife}), or an \sr{object} (e.g. \emph{take out knife}). Later, the CRF unaries use the entire score vectors as features, which leads to noisy predictions.
Namely, \cite{rohrbach13iccv} uses the following formulation for the CRF unaries:
\begin{equation}
\begin{multlined}
E^u(s_n|x_i)=<w_n^u,x_i>,
\end{multlined}
\end{equation}
where $w_n^u$ is a vector of weights between the node-state $s_n$ and the visual attributes. Both $w_n^u$ and $x_i$ have the dimensionality equal to the number of all visual attributes.

Unlike the described method, we train SVM classifiers for visual attributes using their semantic meaning (being a tool, object, etc), e.g. we train different classifiers for a \emph{knife}-\sr{tool} and \emph{knife}-\sr{object}. This allows us to use a score of each node/state classifier directly as a feature for a corresponding unary:
\begin{equation}
\begin{multlined}
E^u(s_n|x_i)=w_n^u x_{i,n}
\end{multlined}
\end{equation}
Here $w_n^u$ is a scalar weight and $x_{i,n}$ is a score of the respective visual classifier. 
Thus we get more discriminative unaries and also reduce the number of parameters of the model (number of connections between node-states and visual features).
The topic node unary $E^u(s_t|v)$ is defined similarly, based on the composite activity recognition features \cite{rohrbach12eccv} as visual descriptors of video $v$.

\subsection{Hand centric features for Object Recognition}
For hand localization we exploit hand appearance
to train an effective hand detector and integrate this detector 
into a upper body pose estimation 
approach. Given the detected hand bounding boxes we densely extract color Sift on 4 channels (RGB+grey) and quantize them in a codebook of size 4000.

\myparagraph{Hand Detector Based on Appearance}
Our hand detector is based on the deformable part models (DPM). 
We aim to differentiate left and right hands as they perform different roles in many activities.
%
Therefore, we dedicate separate DPM components to left and right hands but jointly train them in one detector. 
At test time 
we pick the best scoring hypothesis among the left and right hand components.
%
We found that a rather large number of components is needed to
achieve good detection performance.
DPM components are initialized via k-means clustering of samples by hand
orientation and HOG descriptors.

\myparagraph{Hand Detection Based on Body Pose}
\begin{figure}[t]
\centering
{\includegraphics[width=0.9\linewidth]{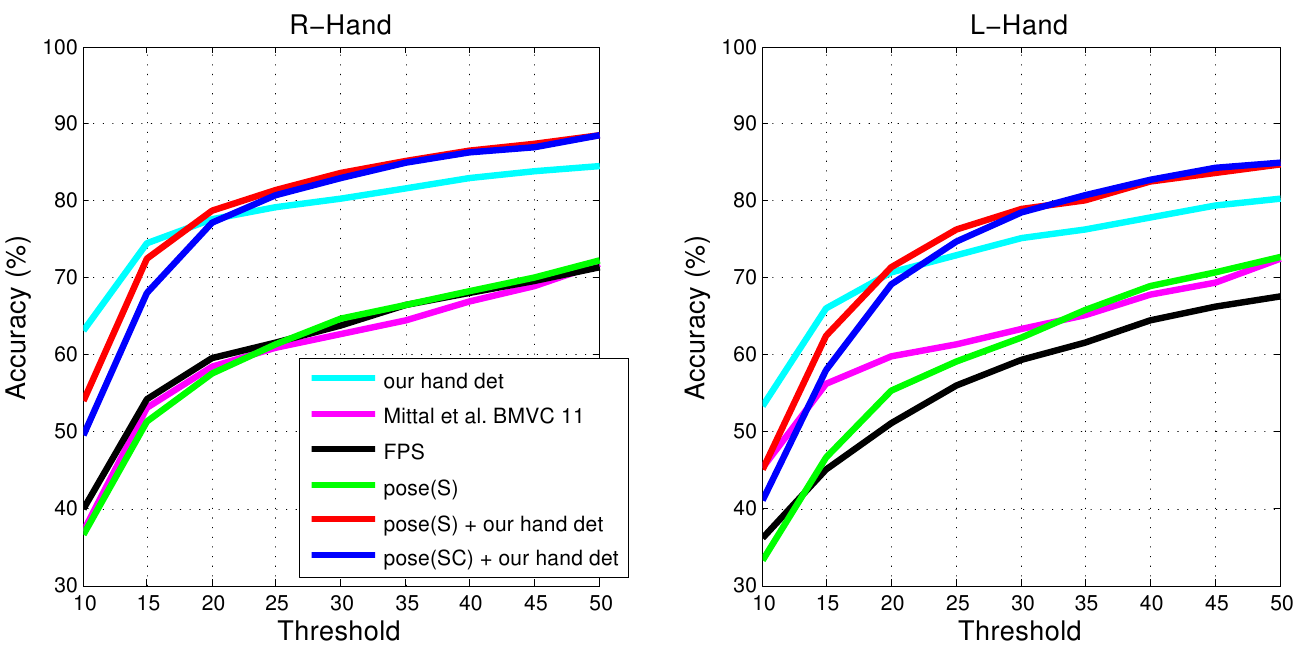}}
\caption{Detection accuracy of right and left hands for
a varying distance from the ground truth position: pose(S) denotes shape context; pose(SC) - shape context and color.}
\label{fig:hand_pose_euc}
\end{figure}
To jointly estimate the hand positions with other body parts we
employ a pictorial structures (PS) model 
%
\cite{andriluka11ijcv}.
The upper body is represented by
10 parts including torso, head, left and right
shoulders, elbows, wrists and hands. 
The model combines a kinematic tree prior for efficient inference 
and body part detectors using shape context features. 
%
We extend this model as follows.
First we train the model using more training data.
Next, we incorporate color features into the part likelihoods by stacking them with
the shape context features. 
%
Finally, we extend the body part detections 
with detection hypotheses for left and right hands based hand detector described above.
Based on the sparse set of non-max suppressed detections we obtain
a dense likelihood map for both hands 
using a Gaussian kernel density estimate.

\myparagraph{Hand Detection Evaluation}
We evaluate our hand detector on the 
``Pose Challenge'' dataset \cite{rohrbach12cvpr} that contains $1277$ test images.
Results are shown in Figure \ref{fig:hand_pose_euc}.
Our hand detector alone significantly improves over the
state-of-the-art FPS approach of \cite{rohrbach12cvpr}.
The performance further improves when hand detectors are integrated in the PS model.
Our detector also significantly improves over the hand detector of
\cite{mittal11bmvc} that in addition to hand appearance also relies on color and
context features. 
%

 

%% file: approachNLG.tex
\vspace{-.1cm}\section{Generating natural descriptions\invisible{ - 0.5 page}}
\label{sec:approach_nlg}
\vspace{-.1cm}
Using a parallel corpus of sentences $z_i$ aligned with a SR $y_i$, i.e.~$(y_i,z_i)$, we adapt SMT techniques \cite{rohrbach13iccv} to generate a novel sentence $z*$ for a SR $y*$.  
\myparagraph{Probabilistic input for SMT}
\begin{figure}[t]
\begin{center}
\includegraphics[width=0.4\textwidth]{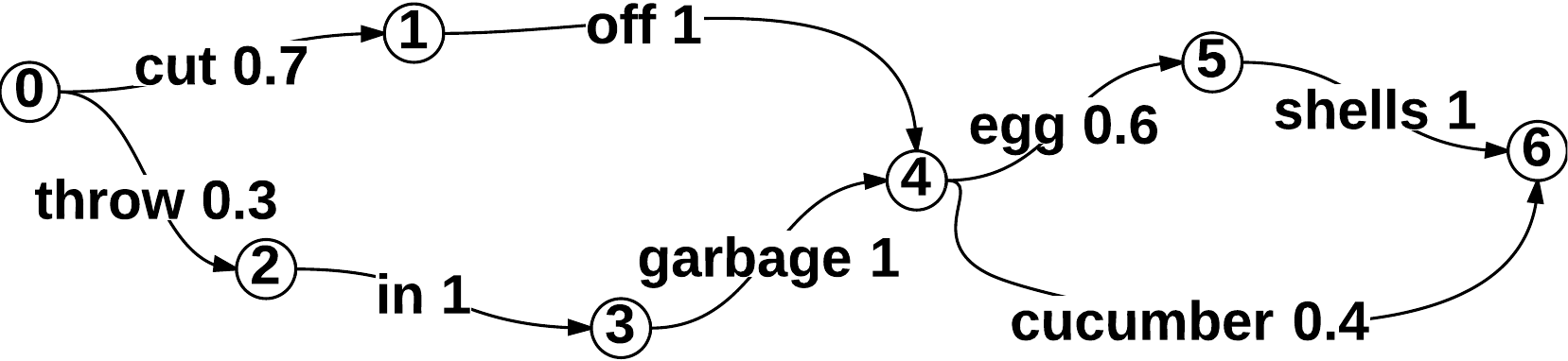}
\caption[labelInTOC]{Encoding probabilistic input for SMT using a word lattice: \sr{$\langle$cut off,egg-shells$\rangle$} has the highest confidence but is unlikely according to the target language model. Thus other candidate paths, e.g. \sr{$\langle$cut off, cucumber$\rangle$} can be a better choice.}
\label{fig:wordLattice}
\end{center}
\figvspace
\figvspace
\end{figure}
While the translation-based approach can achieve performance
comparable with humans on ground truth SRs \cite{rohrbach13iccv}, this does not hold if the SR is noisy.
The approach of \cite{rohrbach13iccv} only takes into account the most probable prediction, the uncertainty found in the SR is not used. However, uncertain input is a known problem for SMT as speech based translation is also based on uncertain input.
The work of \cite{dyer2008generalizing} shows that a probabilistic input encoded in a word lattice can improve the performance of translation by also decoding alternative hypotheses with lower confidence (example see Figure \ref{fig:wordLattice}). 

A \textit{word lattice} is a Directed Acyclic Graph allowing to efficiently decode multiple visual recognition outputs. To construct a word lattice from a set of predicted SRs \sr{$\langle$activity,tool,ingredient,source,target$\rangle$}, we construct a word lattice for each node and then concatenate them. In case that semantic labels are empty in the SRs, we use a symbol \sr{null}$+$\emph{node id} to encode this information in the word lattice.
We  found that providing more flexibility for the alignment model in SMT improves performance, i.e. composite semantic labels such as \sr{cutting-board} are encoded as multiple separate words, e.g. \emph{cutting}, \emph{board}.

SMT combines scores from a phrase-based translation model, a language model, a distortion model and applies word penalties. Word lattice decoding enables us to incorporate confidence scores from the visual recognition. We optimize the model jointly on a development set.

\input{modifyDesc}

%% file: modifyDesc.tex
\myparagraph{Creating cohesive descriptions}
As SMT generates sentences independently for each video segment, the produced multi-sentence descriptions seem more like a `list of sentences' rather than a `text' to readers. Figure \ref{fig:mtOutputDescription} shows an example output of the SMT.
\begin{figure}
\small
\textbf{Output of system}\\
(a) the person got out a cucumber from the fridge\\
(b) the person washed the cucumber\\
(c) the person got out a peeler from the drawer\\
(d) the person got out a knife from the drawer\\
(e) the person peeled the cucumber\\
(f) the person sliced the cucumber\\
\\
\textbf{Post-processed description}\\
(g) A woman got out a cucumber from the fridge.\\
(h) Next, she washed the cucumber.\\
(i) The woman got out a peeler and a knife from the drawer.\\
(j) Finally, she peeled and sliced the cucumber.\\
\figvspace
\caption{Post-processing of descriptions.}
\label{fig:mtOutputDescription}
\figvspace
\end{figure}
\textit{Cohesion} describes the linguistic means which relate sentences on a surface level, and which do not require deep understanding of the text. 
Hence, we automatically post-process the descriptions such that they are more cohesive using a set of domain-independent rules: (1) we fix punctuation and create syntactic parses using the Stanford parser \cite{klein2003accurate}.
(2) We combine adjacent sentences if they have the same verb but different objects. E.g., sentences (c) and (d) in Figure \ref{fig:mtOutputDescription} will be combined to (i).
(3) We combine adjacent sentences if they have the same object but different verbs, e.g., sentences such as (e) and (f) become (j).
(4) The use of referring expressions such as pronouns is a strong cohesive device. As in English, there is no appropriate pronoun for the phrase \textit{the person}, we use gold-standard gender information and replace this phrase by appropriate nouns and pronouns. (5) We insert temporal adverbials such as \textit{next}, \textit{then} and \textit{finally}.

%% file: results.tex
\section{Evaluation\invisible{ - 2 pages}}
\label{sec:results}

We augment the video-description dataset TACoS \cite{regneri13tacl} with short and single sentence descriptions (see Sec. \ref{sec:dataAnalysis}). Following the experimental setup of \cite{rohrbach13iccv}  we use videos and dense trajectory features \cite{wang13ijcv} published by \cite{rohrbach12eccv}; we use the same test split for visual recognition and video description. We preprocess all text data by substituting gender specific identifiers with ``The person'' and, in contrast to \cite{rohrbach13iccv}, transform all sentences to past tense to ensure consistent multi-sentence descriptions.

We evaluate generated text per sentence and per description using BLEU@4, which computes the geometric mean of n-gram word overlaps for n=1,...,4, weighted by a brevity penalty. We also perform human evaluation of produced descriptions asking human subjects to rate readability (without seeing the video), correctness, and relevance (latter two with respect to the video). Readability is evaluated according to the TAC\footnote{www.nist.gov/tac/2011/Summarization/Guided-Summ.2011.guidelines.html} definition which rates the description's grammaticality, non-redundancy, referential clarity, focus, structure and coherence. Correctness is rated per sentences with respect to the video (independent of completeness), we average the score over all sentences per description. Relevance is rated for the full descriptions and judges if the generated description captures the most important events present in the video.
For our segmentation we estimate the best number of initial segment size (60 frames), the similarity measure (cosine), and termination threshold (0.982) on a validation set and fix them for all experiments. 

\subsection{Visual Recognition}
We first evaluate the output of our visual recognition, the SR. We report accuracy of CRF nodes over all labeled ground truth intervals on the test set in Table \ref{tbl:res:vis}. The first line shows the results reported by \cite{rohrbach13iccv}. We notice that the recognition of the handled object (in many cases the ingredient), is the most difficult, achieving only 33.2\% compared to 60.8\% or more for the other nodes. 
This lower performance can be explained by the larger number of states (last line, Table \ref{tbl:res:vis}) and high intra-class variability of the ingredients.
This is in contrast to the importance for verbalization where the activity (second lowest) and handled object are naturally most important (see also Sec. \ref{sec:dataAnalysis}). 

\input{resultTableVisual}

As a first step we add a dish node to the CRF without any features (line 2 in Table \ref{tbl:res:vis}). However, the dish recognition of 8.1\% is too low and enforcing consistency by conditioning on the node prediction confuses the other nodes, leading to a drop in performance for most nodes. Once we add semantic unaries the performance improves for activities by 5.6\% and for objects by 3.9\% compared to  \cite{rohrbach13iccv}. 
Next we improve the dish recognition accuracy by adding more training data
during the CRF training. We use additional videos from the MPII Composite dataset \cite{rohrbach12eccv} that correspond to dishes of the TACoS subset. This data was previously only used for learning attribute/unary classifiers. This leads to an improvement not only for the dish node, but for all nodes (see line 4).
As a next step we add unaries to the dish node. Here we use the features proposed for composite activity recognition \cite{rohrbach12eccv}, training a specific SVM for each state of the dish node. During training and test time we use the ground-truth segmentation for computing the features. Comparing this to the same approach without dish features (line 5 versus 4) improves the dish node significantly from 29\% to 46\%. 
As a last step we add our hand centric color Sift features as second unary for all states from the nodes \sr{tool}, \sr{object}, \sr{source}, and \sr{target}. This leads to a significant improvement for objects of 6.7\% and for \sr{dish} of 9.9\% (line 6 versus 5). 
In comparison to \cite{rohrbach13iccv} we achieve an impressive, overall improvement of 5.3\% for \sr{activity}, 2.5\% for \sr{tool}, 15.9\% for \sr{object}, 0.7\% for \sr{source}, and 2.8\% for \sr{target}. 

 
\subsection{Multi-sentence generation}

We start by using the ground truth intervals provided by TACoS. Results are shown in the upper part of Table~\ref{tbl:res:detailed}. The first line shows the results using the SR and SMT from \cite{rohrbach13iccv} (the best version, learning on predictions), which achieves a BLEU@4 score of 23.2\% when evaluated per sentence. This is an increase from 22.1\% reported by \cite{rohrbach13iccv} due to converting the TACoS corpus to past tense, making it more uniform. The BLEU@4 evaluated per description is 55.7\%\footnote{The BLEU score per description is much higher than per sentence as the the n-grams can be matched to the full descriptions.} and human judges score these descriptions with 2.5 for readability, 3.3 for correctness, and 2.8 for relevance on a scale from 1-5.
Using our improved SR (line 2 in Table \ref{tbl:res:detailed}) consistently improves the quality of the descriptions. Judges rate especially the readability much higher (+0.8) which is due to our increased consistency introduced by the dish node. Also correctness (+0.3) and relevance (+0.2) is rated higher, and the BLEU score improves by 1.9\% and 8.1\%. To estimate the effect of our hand centric features we evaluate our SMT without them, which reaches a BLEU score of 24.1\% (-1.0\%) per sentence  and 61.1\% (-2.7\%) for full descriptions. This indicates that the suggested features have a strong effect not only on the visual recognition performance but also on the quality of our descriptions.

\input{resultTableNlgSentences}

Next, we evaluate the effect of using probabilistic input for SMT in the form of a word lattice (line 3 in Table \ref{tbl:res:detailed}). Again all scores increase. Most notably the BLEU@4 score by 2.3\% and readability by 0.3. While learning on prediction can recover from systematic errors of the visual recognition \cite{rohrbach13iccv}, using probabilistic input for SMT allows to recover from errors made during test time by choosing a less likely configuration according to the visual recognition but more likely according to the language model, e.g. \textit{``The person got out a knife and a cutting board from the pot''} is changed to \textit{``The person took out a pot from the drawer''}. 
We can further improve readability to 3.8 by applying linguistic post-processing to the description (see Sec. \ref{sec:approach_nlg}).
Although we make significant improvements over \cite{rohrbach13iccv}, there is still a gap in comparison to human description, showing the difficulty of the task and the dataset.\footnote{\label{fn:humandesc}The BLEU score for human description is not fully comparable due to one reference less, which typically has a strong effect on the BLEU score.}

After evaluating on the intervals selected by human to describe the video, we now evaluate on our automatic segmentation in the second part of Table \ref{tbl:res:detailed}. We make three observations: first, the relative performance between \cite{rohrbach13iccv}, our SR, and our SR + probabilistic SMT is similar to the one on ground truth intervals. Second, compared to ground truth intervals the performance drops only slightly and our SR + probabilistic SMT still performs better than \cite{rohrbach13iccv} on ground truth intervals. This indicates the good quality of our segmentation. Third, surprisingly the relevance slightly improves for our approaches by 0.1/0.2. This might be due to  our background classifier which removes unimportant segments.

\myparagraph{Qualitative evaluation}
Tables \ref{tbl:res:qual_pos} and \ref{tbl:res:qual_neg} demonstrate the qualitative results of our approach and compare them to human-written descriptions and the output of \cite{rohrbach13iccv}. For the fair comparison we show the output of our system without the post-processing step.
In Table \ref{tbl:res:qual_pos} we illustrate an example when the dish was correctly recognized. Our system produces a consistent multi-sentence description which follows the topic of the video, namely \textit{``Preparing a carrot''}. Unlike ours, the description of \cite{rohrbach13iccv} contains multiple topic changes, which makes it neither readable nor informative for humans.
Table \ref{tbl:res:qual_neg} shows an example where the dish was not correctly identified. Our system predicted \textit{``Preparing orange juice''} instead of \textit{``Juicing a lime''}, confusing the main object of the video. Still, the description is much more relevant than the one of \cite{rohrbach13iccv}, due to its consistency with a similar dish.

\subsection{Multi-level generation}
\label{sec:results:multilevel}

\input{resultTableShort}

Next we evaluate our approach with respect to short (Table \ref{tbl:res:short}) and single sentence (Table \ref{tbl:res:singleSentence}) descriptions. As for detailed descriptions, our improved SR helps to achieve an increase in BLEU and human judgments underlining our above claims. 

The upper part of the Table \ref{tbl:res:short} compares results from \cite{rohrbach13iccv} and our approach on ground truth intervals. 
To produce a short description using our segmentation, we select top 3 relevant segments, as described earlier (Sec. \ref{sec:approach_system}). We decide for 3 segments as the average length of short descriptions in the corpus is 3.5 sentences. We compare different approaches of producing short descriptions. First line shows the result of extracting sentences from the detailed description generated by model of \cite{rohrbach13iccv} trained on TACoS. Second line corresponds to short description generated by \cite{rohrbach13iccv} trained on the short descriptions. Similarly next two lines correspond to extracted and generated short descriptions produced by our system. In both cases we observe that language models specifically trained on the short descriptions perform better. This supports our hypothesis that for the best performance we need to learn a language model for a desired level of detail. Interestingly, the descriptions produced on our segmentation got higher or similar human judgment scores than on ground truth Short Desc. intervals. This shows, that our method to select relevant segments indeed captures the most important events of the video. 

\input{resultTableSingleSentence}

Finally, Table \ref{tbl:res:singleSentence} shows the results for the single sentence description generation. The first line of the table shows the result of the retrieval based on the predicted dish. We select a sentences that describes the closest training video of the same dish, using the dish unary features. This results in a BLEU@4 score 23.3\%, which is far below 48.8\% for human descriptions. 
The last four lines compare the extractively produced descriptions. Here we have the same competing methods as in Table \ref{tbl:res:short}; we extract a single sentence either from the detailed or short description. The best performance is achieved by our model trained on the short descriptions (last line). Interestingly it significantly outperforms the retrieval-based approach, due to more accurate recognition of activities/objects present in test video. 

\input{resultTablesQuality}

%% file: resultTableVisual.tex
\newcommand{\midrulecrf}{\cmidrule(lr){1-1}  \cmidrule(lr){2-6}\cmidrule(lr){7-7}}
\newcommand{\midrulecrfsum}{\cmidrule(lr){1-1}   \cmidrule(lr){2-7}}

\newcommand{\ps}{\hphantom{+ }}

\begin{table}
\center
\small
\setlength{\tabcolsep}{5pt}
\begin{tabular}{lr@{\ \ \ }r@{\ \ \ }r@{\ \ }r@{\ \ }rr}
\toprule
Approach & {acti.} & {tool}  & {obj.} & {source} & {target}
& {dish} \\ 
\midrulecrf
CRF \cite{rohrbach13iccv}& 60.8 & 82.0 & 33.2 & 76.0 & 74.9 &-\\
\midrulecrf
CRF + dish consistency     & 60.4 & 83.0 & 30.0 & 70.7 & 69.6& 8.1 \\
+ Semantic unaries 	 & \textbf{66.4} & 82.3 & 37.1 & 77.0 & 77.0 & 12.7 \\
\ps+ Data (CRF training) & \textbf{66.4} & 83.4 & 41.0 & \textbf{78.1} & 77.4 & 29.0 \\
\ps\ps+ Dish unaries& 65.7 & 83.0 & 42.4 & 76.7 & 76.3 & 46.3 \\
\ps\ps\ps+ Hand centric&66.1& \textbf{84.5} & \textbf{49.1} & 76.7 & \textbf{77.7} & \textbf{56.2} \\
\midrule
number of states & 66 & 43 & 109 & 51 & 35 & 26\\
\bottomrule 
\end{tabular}
\caption{Visual recognition of SR, accuracy in \%.}
\label{tbl:crfnodes}
\label{tbl:res:vis}
\figvspace
\end{table}

%% file: resultTableNlgSentences.tex
\newcommand{\midruleResTrans}{\cmidrule(lr){1-1}  \cmidrule(lr){2-2} \cmidrule(lr){3-6}}
\begin{table}[t]
\center
\small
\begin{tabular}{l  r r@{\ \ } r@{\ \ }r@{\ \ }r}
\toprule
\multicolumn{2}{r}{\scriptsize PER SENTENCE} & \multicolumn{4}{l}{\scriptsize PER DESCRIPTION} \\
&  &  \multicolumn{1}{c}{} & \multicolumn{3}{c}{\footnotesize Human judgments} \\
Approach & BLEU  &  BLEU &  Read. & Corr. & Rel.\\
\midruleResTrans
\multicolumn{3}{l}{\textbf{On TACoS gt intervals}}\\
SMT \cite{rohrbach13iccv} & 23.2 & 55.7 &  2.5 & 3.3 & 2.8 \\
SMT (our SR) & 25.1 &  63.8 &  3.3 & 3.6 &  3.0  \\
+ probabilistic input & {27.5} & 66.1  & 3.6 & 3.7 &3.1 \\
Human descriptions & 36.0\textsuperscript{\ref{fn:humandesc}} & 63.6\textsuperscript{\ref{fn:humandesc}}  & 4.4 & 4.9& 4.8  \\
\midruleResTrans
\multicolumn{2}{l}{\textbf{On our segmentation}}\\
SMT \cite{rohrbach13iccv} & - & 43.9 &  2.4  & 2.9& 2.7  \\
SMT (our SR)  & - & 55.3&    2.7 &  3.4&  3.2  \\
+ probabilistic input & - & 55.8 &  3.2 & 3.7& 3.3  \\
\bottomrule
\end{tabular}
\caption{Evaluating \textbf{detailed descriptions}. BLEU@4 in
\%. Human judgments from 1-5, where 5 is best.}
\label{tbl:res:detailed}
\figvspace
\end{table}

%% file: resultTableShort.tex
\renewcommand{\midruleResTrans}{\cmidrule(lr){1-1}  \cmidrule(lr){2-2} \cmidrule(lr){3-6}}
\begin{table}[t]
\center
\small
\begin{tabular}{l  r r@{\ \ }r @{\ \ }r@{\ \ }r}
\toprule
 \multicolumn{2}{r}{\scriptsize PER SENTENCE} & \multicolumn{4}{l}{\scriptsize PER DESCRIPTION} \\
 &  &  \multicolumn{1}{c}{} & \multicolumn{3}{c}{\footnotesize Human judgments} \\
Approach & BLEU  &  BLEU &  Read. & Corr. & Rel.\\
\midruleResTrans
\multicolumn{5}{l}{\textbf{On Short Desc. intervals}}\\
SMT \cite{rohrbach13iccv} &  19.1 & 43.8  & 3.1 & 2.9 & 2.5 \\
SMT (our SR) + prob. & 22.5& 54.4 & 3.6 & 3.5& 3.3 \\
Human descriptions & - & 57.9\textsuperscript{\ref{fn:humandesc}}  & 4.7 & 4.8 & 4.7 \\ 
\midruleResTrans
\multicolumn{5}{l}{\textbf{On our segmentation (top 3)}}\\
\multicolumn{2}{l}{SMT \cite{rohrbach13iccv} (TACoS) }    & 38.7  & -  & -  & - \\
\multicolumn{2}{l}{SMT \cite{rohrbach13iccv} (Short Desc)}     &41.6 &   3.6 & 2.9 & 2.5 \\
\multicolumn{2}{l}{SMT (our) + prob. (TACoS)}    & 47.2 &  -  & -  & - \\

\multicolumn{2}{l}{SMT (our)  + prob.  (Short Desc)}    & 56.9 & 4.2 & 3.5 & 3.3 \\

\bottomrule
\end{tabular}
\vspace{-0.15cm}
\caption{Evaluating \textbf{short descriptions}. BLEU@4 in
\%. Human judgments from 1-5, where 5 is best.}
\label{tbl:res:short}
\end{table}

%% file: resultTableSingleSentence.tex
\renewcommand{\midruleResTrans}{\cmidrule(lr){1-1} \cmidrule(lr){2-5}}
 \begin{table}[t]
\center
\small
\begin{tabular}{l  r@{\ \ }r @{\ \ }r@{\ \ }r}
\toprule
  & \multicolumn{4}{l}{\scriptsize PER DESCRIPTION} \\
  &  \multicolumn{1}{c}{} & \multicolumn{3}{c}{\footnotesize Human judgments} \\
Approach & BLEU &  Read. & Corr. & Rel.\\
\midruleResTrans
\multicolumn{2}{l}{\textbf{On full video}}\\
  Retrieval (Nearest neighbor) & 23.3 &  4.6  & 3.3 & 3.3 \\
 Human descriptions & 48.8 &  4.7 & 4.5&    4.5 \\
\midruleResTrans
\multicolumn{2}{l}{\textbf{On our segmentation (top 1)}}\\
\multicolumn{1}{l}{SMT \cite{rohrbach13iccv} (TACoS) } & 18.3 &   -  & -  & - \\
\multicolumn{1}{l}{SMT \cite{rohrbach13iccv} (Short Desc)}  & 25.5 & 4.6 & 2.5& 2.5 \\
\multicolumn{1}{l}{SMT (our) + prob. (TACoS)}  &  25.6&   -  & -  & - \\
\multicolumn{1}{l}{SMT (our)  + prob.  (Short Desc)}   & 28.5 &  4.7 & 3.5 & 3.5 \\
\bottomrule
\end{tabular}
\caption{Evaluating  \textbf{single sentence descriptions}. BLEU@4~in~
\%. Human judgments from 1-5, 5 is best.}
\label{tbl:res:singleSentence}
\figvspace
\end{table}

%% file: resultTablesQuality.tex
\begin{table*}
\centering
\small
\begin{tabular}{p{4.5cm} p{6cm} p{6cm}}
\hline
Human Description & SMT (our SR) + probabilistic input & SMT \cite{rohrbach13iccv}\\
\hline
The person took out carrots. & The person took out a carrot from the refrigerator \textbf{and placed it on the cutting board}. & The person took out carrots from the fridge. \\
The person took out knife. & The person took out \textbf{a cutting board and} knife. & The person took out a knife from the drawer \textbf{and a cutting board from a drawer}. \\
The person took out cutting board. & The person took out a cutting board from the drawer. & The person took out a cutting board from the drawer. \\
The person took out plate. & The person got out a plate from the cabinet. & The person took out \textbf{pomegranate and bowl} from the cabinet.  \\
The person washed carrot. &	The person washed the carrot. &	The person \textbf{grated half of the cucumber}. \\
The person took out peeler. & The person got out a peeler from the drawer. & The person took out \textbf{a cutting board a knife and a carrot}. \\
The person peeled carrot. & The person peeled the carrot \textbf{in the sink}. & The person \textbf{cut the orange grater}. \\
The person threw away peels. & 	The person threw away the peels into the wastebin. & The person threw away the peel.  \\
The person chopped up carrot. & The person \textit{put} the carrots on the cutting board. & The person cut up \textbf{the beans}. \\
The person put back extra carrots. & The person \textbf{took out} a carrot \textbf{from} the fridge. & The person \textbf{took out a knife from the person took out} a carrot.  \\
\hline
\end{tabular}
\caption{Qualitative comparison of the human descriptions, SMT (our SR) + probabilistic input and SMT \cite{rohrbach13iccv}. Example where the dish was correctly recognized. Mistakes are marked in bold. Correct but not the most relevant information is marked in italic.}
\label{tbl:res:qual_pos}
\end{table*}

\begin{table*}
\centering
\small
\begin{tabular}{p{4.5cm} p{6cm} p{6cm}}
\hline
Human Description & SMT (our SR) + probabilistic input & SMT \cite{rohrbach13iccv}\\
\hline
The person took the lemon squeezer out of the drawer. & The person took out a \textbf{cutting board} from the drawer. & The person took out \textbf{a cutting board} from the drawer \textbf{on the cutting board}. \\
The person took a lime out of the refrigerator. &	The person took out an \textbf{orange} from the fridge. & The person got out the \textbf{broccoli} from the fridge. \\
The person took a knife out of the drawer. & The person got out a knife from the drawer. & The person took out a knife from the drawer. \\
The person took a round chopping board out of the drawer. & The person took out a cutting board from the drawer. & 	The person took out a cutting board from the drawer \textbf{on the cutting board}.  \\
The person cut the lime in half. &	The person cut the \textbf{orange} in half. &	The person cut the \textbf{avocado in on the plate}. \\
The person squeezed the juice out of both lime halves. & The person juiced the \textbf{orange}. & The person \textbf{cut the orange} on juicer. \\
The person washed her hands. & The person washed her hands. & 	The person \textbf{rinsed the chili}. \\
\hline
\end{tabular}
\caption{Qualitative comparison of the human descriptions, SMT (our SR) + probabilistic input and SMT \cite{rohrbach13iccv}. Example where the dish was wrongly recognized. Mistakes are marked in bold.}
\label{tbl:res:qual_neg}
\end{table*}

%% file: conclusion.tex
\section{Conclusion\invisible{ - 1 page incl. refs}}
 \label{sec:conclusion}

This work addresses the challenging task of coherent multi-sentence video descriptions. We show that inferring the high level topic helps to ensure consistency across sentences.
We also show that hand centric features help to improve the visual recognition of the most challenging semantic category, namely manipulated objects, which leads to consecutive improvements of produced descriptions. 
Another finding of our work is that using probabilistic input in SMT improves the quality of sentences produced for individual video snippets.

We also address the so far unexplored task of producing video descriptions at multiple levels of detail. We collected and analyzed a corpus of human descriptions of three levels of detail in order to find out how the language changes across these levels. We found that with decreasing length of description, the verbalized information is being `compressed' according to the topic of the video. Based on this we propose a method to extract most relevant segments of the video. 
We also show that using a language model targeted at the type of description improves over using one learned from descriptions of another level of detail.
  

